**AI-Enabled Knowledge Sharing for Enhanced Collaboration and Decision-Making in Non-Profit Healthcare Organizations: A Scoping Review Protocol**

**Maurice Ongala Ouma[1], Ruth Kiraka[2], Jyoti Choudrie[3], Javan Solomon Okello[4]**

[1] Strathmore University, Kenya
[2] Strathmore University, Kenya
[3] University of Hertfordshire, United Kingdom
[4] University of Kwa Zulu Natal, South Africa

[1] maurice.ouma@strathmore.edu
[2] rkiraka@strathmore.edu
[3] j.choudrie@herts.ac.uk
[4] javanokello@gmail.com

**Abstract**

This protocol outlines a scoping review designed to systematically map the existing body of evidence on AI-enabled knowledge sharing in resource-limited non-profit healthcare organizations. The review aims to investigate how such technologies enhance collaboration and decision-making, particularly in the context of reduced external support following the cessation of USAID operations. Guided by three theoretical frameworks namely, the Resource-Based View, Dynamic Capabilities Theory, and Absorptive Capacity Theory, this study will explore the dual role of AI as a strategic resource and an enabler of organizational learning and agility. The protocol details a rigorous methodological approach based on PRISMA-ScR guidelines, encompassing a systematic search strategy across multiple databases, inclusion and exclusion criteria, and a structured data extraction process. By integrating theoretical insights with empirical evidence, this scoping review seeks to identify critical gaps in the literature and inform the design of effective, resource-optimized AI solutions in non-profit healthcare settings.

1. **Introduction**

**1.1 Background and Rationale**

The integration of Artificial Intelligence (AI) into healthcare has expanded significantly in the recent past, offering transformative solutions to challenges in diagnostics, predictive analytics, telemedicine, and supply chain optimization. AI-driven technologies have demonstrated potential in addressing shortages of healthcare professionals, enhancing disease surveillance, and improving healthcare delivery in underserved regions (1,2). For instance, AI-powered diagnostic tools for tuberculosis and radiology have achieved accuracy comparable to expert clinicians, thus augmenting healthcare access in remote areas with limited specialist (1) Additionally, AI-based predictive analytics has strengthened disease surveillance by enabling early outbreak detection and response, as evidenced during the COVID-19 pandemic (3–5)

Despite these advancements, challenges persist in implementing AI solutions in low-resource settings, necessitating improved collaboration and decision-making frameworks. Limited infrastructure, lack of local datasets for AI training, and ethical concerns regarding data privacy and bias hinder large-scale adoption (4,5) Successful integration of AI in healthcare requires a coordinated effort among governments, healthcare providers, and technology developers to establish robust regulatory frameworks, invest in digital infrastructure, and provide training for healthcare workers (6). Moreover, equitable AI deployment must consider ethical principles, ensuring that AI tools are transparent, contextually relevant, and do not exacerbate health disparities (5)

Addressing these challenges necessitates interdisciplinary and cross-sectoral collaboration to develop AI solutions that are scalable, sustainable, and aligned with public health priorities in resource-constrained environments. With proper governance and strategic investment, AI can bridge healthcare access gaps, optimize clinical workflows, and enhance healthcare equity worldwide (7)

**1.2 Problem Statement and Objectives**

The rapid evolution of AI technologies has led to an increased deployment of AI-enabled knowledge-sharing systems in various sectors, including healthcare(8). However, despite the promise of these technologies, there remains a significant gap in understanding how such systems specifically impact non-profit healthcare organizations. This gap is particularly pronounced in resource-limited settings, where external funding and support, such as that previously provided by USAID, have been reduced or withdrawn (9–11). As a result, non-profit healthcare organizations are under increased pressure to optimize their internal resources and adopt innovative approaches to maintain effective collaboration and decision-making.

The primary purpose of this scoping review is to systematically map the existing body of evidence on AI-enabled knowledge sharing in non-profit healthcare organizations. Additionally, the review aims to integrate theoretical insights drawn from the Resource-Based View, Dynamic Capabilities Theory, and Absorptive Capacity Theory, to provide a comprehensive understanding of how these technologies function as strategic resources. By doing so, this study seeks to identify critical gaps in the literature and propose directions for future research that can inform the design of more effective, resource-optimized AI solutions in the post-USAID support era.

### 1.3 Primary Research Question

To address this purpose, the following research question has been formed:

*What is the scope of the existing literature on AI-enabled knowledge sharing for enhancing collaboration and decision-making in resource-limited non-profit healthcare organizations?*

### 1.4 Subsidiary Research Questions

i. How do AI-enabled knowledge-sharing systems function as strategic resources in these organizations?
ii. In what ways are theoretical frameworks—namely, the Resource-Based View, Dynamic Capabilities Theory, and Absorptive Capacity Theory —applied to understand the impact of these systems?
iii. What gaps and challenges exist in the literature regarding the implementation and effectiveness of AI-enabled knowledge sharing in resource-constrained healthcare settings?

## 1.5 Rationale for the Scoping Review

Given the rapid advancements in AI and its multifaceted applications across healthcare and organizational knowledge management, the scoping review approach allows for a systematic synthesis of the extant research, enabling the identification of key themes, critical gaps, and future research directions (12). Unlike systematic reviews that focus on narrowly defined research questions, scoping reviews embrace the breadth of a fragmented literature, integrating studies from computer science, healthcare, organizational studies, and non-profit management (13)

In this context, where AI applications range from decision support systems to collaborative knowledge-sharing platforms, a scoping review facilitates an extensive mapping of the extent, range, and nature of research, clarifying how these technologies enhance collaboration, decision-making, and operational efficiency (14). Interdisciplinary research often grapples with inconsistent terminologies, diverse methodologies, and varying theoretical underpinnings(15). A scoping review is uniquely positioned to integrate insights from multiple theoretical frameworks namely, the Resource-Based View, and Dynamic Capabilities Theory, thus elucidating AI's dual role as a strategic resource and an enabler of organizational learning (13)

> **Commented [JC1]:** Why these theories

## 2 Theoretical Framework

### 2.1 Resource-Based View (RBV)

The Resource-Based View (RBV) posits that organizations achieve sustained competitive advantage by leveraging unique, valuable, rare, and inimitable resources (16). In non-profit healthcare organizations, AI-driven tools and knowledge-sharing platforms constitute strategic resources that enhance healthcare project performance by optimizing decision-making, improving knowledge dissemination, and fostering operational efficiency (17). AI-powered analytics enable organizations to process vast amounts of healthcare data, transforming raw information into actionable insights that inform patient care, resource allocation, and policy formulation (18).

Furthermore, AI-driven knowledge-sharing platforms facilitate real-time collaboration among healthcare providers, reducing information asymmetry and improving organizational learning, ultimately strengthening healthcare delivery in resource-limited settings (19)

**2.2 Dynamic Capabilities Theory**

Dynamic Capabilities Theory (DCT) emphasizes an organization's ability to sense, seize, and transform opportunities in volatile environments (20). AI enhances these capabilities by enabling healthcare organizations to detect emerging health trends (sensing), develop innovative interventions (seizing), and reconfigure resources for improved service delivery (transforming) (17). In non-profit healthcare settings, AI-driven predictive analytics and automated decision-support systems enhance agility, allowing organizations to rapidly adapt to changes in disease epidemiology, and patient needs (21,22). AI-enabled systems also foster process innovation by streamlining workflows, reducing administrative burdens, and optimizing resource utilization, ultimately improving healthcare outcomes in non-profit contexts (23).

**2.3 Absorptive Capacity Theory**

Absorptive Capacity Theory (ACT) explains how organizations acquire, assimilate, transform, and exploit new knowledge (24). In healthcare, AI-generated knowledge is only impactful if organizations develop absorptive capacity to integrate and utilize it effectively (25) AI-driven tools enhance organizational learning by providing data-driven insights that support evidence-based decision-making, thereby improving clinical effectiveness and operational efficiency (26). Additionally, non-profit healthcare organizations benefit from AI-powered knowledge-sharing mechanisms that promote stakeholder engagement, ensuring that frontline workers, policymakers, and community health providers can collaboratively leverage AI-derived insights to enhance patient care and public health interventions (27)

**2.4 Theoretical Integration**

The integration of RBV, DCT, and ACT provides a comprehensive theoretical framework for examining AI-enabled knowledge sharing in non-profit healthcare organizations. By synthesizing these perspectives, this scoping review systematically maps how AI functions as a strategic enabler of knowledge dissemination and decision-making in non-profit healthcare settings, identifying gaps and future research directions for optimizing AI deployment in the post-USAID support era.

**3.0   Methodology**

**3.1   Review Design**

This scoping review will be conducted in accordance with the PRISMA-ScR (Preferred Reporting Items for Systematic Reviews and Meta-Analyses extension for Scoping Reviews) guidelines to ensure methodological rigor, transparency, and reproducibility (28). The review design is chosen to systematically map the interdisciplinary literature on AI-enabled knowledge sharing in resource-limited non-profit healthcare organizations. Given the diverse study designs and theoretical perspectives spanning computer science, healthcare, and organizational studies, a scoping review is uniquely positioned to capture the breadth and complexity of this research field.

**3.2   Search Strategy**

A comprehensive search will be executed across multiple electronic databases, including PubMed, Scopus, IEEE Xplore and Google Scholar. PubMed provides authoritative biomedical and health sciences literature, making it essential for research on AI in healthcare, knowledge sharing, and non-profit health interventions. Scopus offers multidisciplinary coverage, ensuring access to peer-reviewed literature on AI, healthcare management, and organizational theories. IEEE Xplore is the leading database for engineering and AI research, offering insights into machine learning, predictive analytics, and emerging technologies in healthcare. Google Scholar broadens access to interdisciplinary studies, including grey literature and conference papers, enhancing retrieval of research at the intersection of AI, knowledge management, and non-profit healthcare.

The search strategy will integrate a combination of keywords and controlled vocabulary related to artificial intelligence (e.g., "AI," "machine learning," "predictive analytics"), knowledge sharing (e.g., "knowledge management," "collaborative platforms"), non-profit healthcare (e.g., "resource-limited healthcare," "public health non-profit"), and the relevant theoretical frameworks (e.g., "Resource-Based View," "Dynamic Capabilities Theory," "Absorptive Capacity").

The search will be restricted to articles published in English from 2019 to the present, covering empirical studies, reviews, conceptual papers, and theoretical analyses from relevant disciplines, including artificial intelligence, healthcare management, public health, organizational studies, non-profit management, and knowledge management. These studies will align with the research objectives of mapping AI-enabled knowledge sharing in non-profit healthcare organizations and integrating insights from relevant theoretical frameworks

**3.3     Inclusion and Exclusion Criteria**

Studies will be selected based on their focus on AI-enabled knowledge sharing within non-profit healthcare settings, explicitly addressing collaboration, decision-making, or the application of the theoretical frameworks. Articles focusing solely on for-profit contexts, non-healthcare sectors, or those that lack an explicit connection to AI-enabled interventions will be excluded.

**3.4     Study Selection Process**

The study selection process will involve an initial screening of titles and abstracts by two independent reviewers, followed by a full-text review to confirm eligibility. Discrepancies will be resolved through consensus or consultation with a third reviewer. Reference management software (EndNote) and systematic review tools such as Rayyan and Elicit will be employed to streamline this process.

**3.5     Data Extraction and Charting**

Data extraction will be conducted using a standardized form designed to capture study characteristics, including authorship, publication year, geographical context, study design, and details of the non-profit healthcare setting. Information regarding the AI methodologies employed, the application of theoretical frameworks, and reported outcomes related to collaboration and decision-making will also be systematically charted. Studies will be coded according to theoretical perspective and thematic content to facilitate a structured synthesis of the literature.

**3.6     Quality Assessment**

Although scoping reviews typically do not mandate formal quality appraisal, an assessment of methodological rigor will be conducted to provide contextual insight into the reliability and validity of the findings.

**4.0     Data Synthesis and Analysis**

Following data extraction, a thematic analysis will be undertaken to systematically identify and synthesize patterns across the literature. The review will explore thematic domains including AI

Implementation and Resource Utilization, Organizational Agility and Capability Building, Knowledge Acquisition and Absorptive Capacity, and Knowledge Management Strategies. Each theme will be aligned with one or more of the theoretical perspectives such as Resource-Based View, Dynamic Capabilities Theory, and Absorptive Capacity Theory. Tables, charts, and conceptual maps will be employed to visually represent the integration of theory and empirical findings. Additionally, a detailed framework will be applied to map each study against the four theoretical lenses, highlighting consistencies, discrepancies, and research gaps that can direct future inquiries.

**5.0     Ethical Considerations**

This review will adhere to strict ethical standards throughout the research process. Although the study synthesizes published data, the review will ensure transparency in data handling, analysis, and reporting. It will accurately represent original findings, maintain data privacy, and address any potential conflicts in the literature. The ethical review process will follow established institutional guidelines for research synthesis, ensuring that all analyses are conducted responsibly and impartially.

**6.0     Dissemination Plan**

The findings of this review will be disseminated in a high-impact, peer-reviewed journal BMJ Open and presented at HAIC2025, the Symposium on Human-AI Collaboration. This conference emphasizes user-centered AI, explainability, human-in-the-loop approaches, and ethical considerations which are areas directly relevant to the study. Additionally, the results will be shared with policymakers, healthcare practitioners, and AI developers through targeted workshops, policy briefs, interdisciplinary exchanges and online media platforms.

Commented [JC2]: Such as?

**Appendix**

**Search Strategy for PubMed**

(("Organizations, Nonprofit"[Mesh] OR "non-profit healthcare"[tiab] OR "nonprofit healthcare"[tiab] OR "not-for-profit healthcare"[tiab] OR "charitable healthcare"[tiab] OR "voluntary health sector"[tiab] OR "third sector health services"[tiab] OR "non-governmental healthcare"[tiab] OR "nonprofit hospitals"[tiab] OR "not-for-profit hospitals"[tiab] OR "charity hospitals"[tiab] OR "community health organizations"[tiab] OR "faith-based healthcare"[tiab] OR "volunteer health services"[tiab] OR "nonprofit health organizations"[tiab] OR "nonprofit health systems"[tiab] OR "nonprofit medical institutions"[tiab] OR "nonprofit clinics"[tiab] OR "community-based healthcare"[tiab])
AND ("Artificial Intelligence"[Mesh] OR "artificial intelligence"[tiab] OR "AI"[tiab] OR "machine intelligence"[tiab] OR "computational intelligence"[tiab] OR "automated decision-making"[tiab] OR "intelligent systems"[tiab] OR "intelligent automation"[tiab] OR "cognitive computing"[tiab] OR "AI-driven technology"[tiab] OR "automated reasoning"[tiab] OR "machine learning"[tiab] OR "deep learning"[tiab] OR "neural networks"[tiab] OR "supervised learning"[tiab] OR "unsupervised learning"[tiab] OR "reinforcement learning"[tiab] OR "predictive analytics"[tiab] OR "AI-based decision support"[tiab] OR "natural language understanding"[tiab] OR "speech recognition"[tiab] OR "text mining"[tiab] OR "computational linguistics"[tiab] OR "computer vision"[tiab] OR "image recognition"[tiab] OR "facial recognition"[tiab] OR "pattern recognition"[tiab] OR "robotic process automation"[tiab] OR "autonomous systems"[tiab] OR "intelligent agents"[tiab] OR "smart automation"[tiab] OR "AI in healthcare"[tiab] OR "AI-driven diagnostics"[tiab] OR "clinical decision support systems"[tiab] OR "AI-assisted radiology"[tiab] OR "AI in drug discovery"[tiab] OR "big data analytics"[tiab] OR "data mining"[tiab] OR "predictive modeling"[tiab] OR "knowledge discovery in databases"[tiab] OR "algorithmic decision-making"[tiab] OR "AI ethics"[tiab] OR "algorithmic bias"[tiab] OR "explainable AI"[tiab] OR "fairness in AI"[tiab])
AND ("Delivery of Health Care"[Mesh] OR "healthcare"[tiab] OR "health care"[tiab])
AND ("2019/01/01"[PDAT]:"3000/12/31"[PDAT])
AND (English[lang])